\documentclass[runningheads]{llncs}
\usepackage[T1]{fontenc}
\usepackage{booktabs}
\usepackage{amssymb}
\usepackage{hyperref}
\usepackage{amsmath}
\usepackage{multirow}
\usepackage{graphicx,verbatim}

\begin{document}
\title{TextBraTS: Text-Guided Volumetric Brain Tumor Segmentation with Innovative Dataset Development and Fusion Module Exploration}
\titlerunning{TextBraTS: Dataset Innovation and Fusion Module Exploration}
%
\begin{comment}  %% Removed for anonymized MICCAI 2025 submission
\author{First Author\inst{1}\orcidID{0000-1111-2222-3333} \and
Second Author\inst{2,3}\orcidID{1111-2222-3333-4444} \and
Third Author\inst{3}\orcidID{2222--3333-4444-5555}}
%
\authorrunning{F. Author et al.}
\institute{Princeton University, Princeton NJ 08544, USA \and
Springer Heidelberg, Tiergartenstr. 17, 69121 Heidelberg, Germany
\email{lncs@springer.com}\\
\url{http://www.springer.com/gp/computer-science/lncs} \and
ABC Institute, Rupert-Karls-University Heidelberg, Heidelberg, Germany\\
\email{\{abc,lncs\}@uni-heidelberg.de}}

\end{comment}

\author{
  Xiaoyu Shi\inst{1} \and
  Rahul Kumar Jain\inst{1} \and
  Yinhao Li\inst{1} \and
  Ruibo Hou\inst{1} \and
  Jingliang Cheng\inst{2} \and
  Jie Bai\inst{2} \and
  Guohua Zhao\inst{2} \and
  Lanfen Lin\inst{3} \and
  Rui Xu\inst{4} \and
  Yen-wei Chen\inst{1}\thanks{is the corresponding author}
}
\authorrunning{Shi et al.}
\institute{
  Ritsumeikan University, Osaka, Japan\\
  \email{chen@is.ritsumei.ac.jp}
  \and
  Department of Magnetic Resonance Imaging, the First Affiliated Hospital of Zhengzhou University, Zhengzhou, China\\
  \and
  Zhejiang University, Hangzhou, China\\
  \and
  Dalian University of Technology, Dalian, China\\
}

\maketitle           % typeset the header of the contribution
\begin{abstract}
Deep learning has demonstrated remarkable success in medical image segmentation and computer-aided diagnosis. In particular, numerous advanced methods have achieved state-of-the-art performance in brain tumor segmentation from MRI scans. While recent studies in other medical imaging domains have revealed that integrating textual reports with visual data can enhance segmentation accuracy, the field of brain tumor analysis lacks a comprehensive dataset that combines radiological images with corresponding textual annotations. This limitation has hindered the exploration of multimodal approaches that leverage both imaging and textual data. To bridge this critical gap, we introduce the TextBraTS dataset, the first publicly available volume-level multimodal dataset that contains paired MRI volumes and rich textual annotations, derived from the widely adopted BraTS2020 benchmark. Building upon this novel dataset, we propose a novel baseline framework for text-guided volumetric medical image segmentation. Through extensive experiments with various text-image fusion strategies and templated text formulations, our approach demonstrates significant improvements in brain tumor segmentation accuracy, offering valuable insights into effective multimodal integration techniques. Our dataset, implementation code, and pre-trained models are publicly available at \href{https://github.com/Jupitern52/TextBraTS}{TextBraTS}.

% Authors must provide keywords and are not allowed to remove this Keyword section.

\end{abstract}
\section{Introduction}

\begin{figure}
    \centering
    \includegraphics[width=0.6\linewidth]{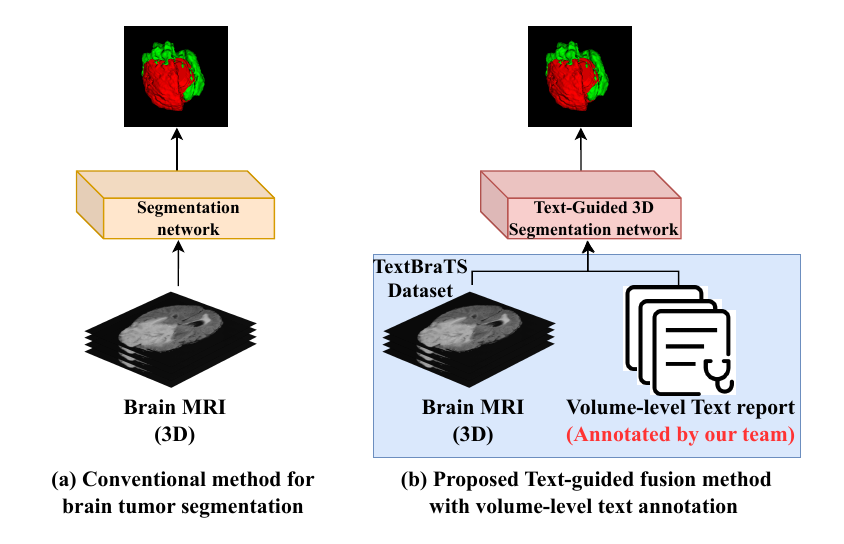}
    \caption{Comparison between conventional brain tumor segmentation methods and our proposed method: (a) Segmentation network for tumor segmentation with image modality input only; (b) Our proposed text-guided multimodal segmentation network with a novel volume-level text report dataset.      }
    \label{overview}
\end{figure}

Brain tumors are highly dangerous malignancies with low survival rates  \cite{louis_2021_2021}. In clinical practice, brain tumor diagnosis and treatment are typically guided by the analysis of tumor location, shape, and morphology using brain MRI scans \cite{bakas_advancing_2017}. Tumor segmentation plays a critical role in computer-aided diagnosis systems, as the accurate delineation of tumor regions is essential for downstream tasks such as tumor grading and genetic marker prediction \cite{shi2023intra}\cite{zhang2023idh}. With advancements in deep learning, several slice-based methods \cite{shi2025acm}\cite{zhao2018deep} and volume-based methods  \cite{crimi_swin_2022}\cite{crimi_brain_2022}\cite{isensee_nnu-net_2021}\cite{navab_u-net_2015}\cite{wang_nestedformer_2022} have been developed to segment tumor regions based on MRI scans. However, a significant limitation of these approaches is their reliance on imaging data alone, without incorporating complementary multimodal information. In practical diagnosis, doctors often integrate radiological reports with imaging data to achieve a more comprehensive diagnostic evaluation \cite{reiner_radiology_2007}. The absence of multimodal integration in existing methods underscores a critical gap in their applicability and effectiveness. Furthermore, the majority of publicly available brain tumor segmentation datasets are restricted to a single imaging modality. This limitation, which impedes the development of advanced methodologies that could leverage the synergistic potential of integrating textual radiological reports with imaging data.
In other medical image segmentation domains, several methods have integrated textual reports with imaging data to achieve segmentation performance superior to that of single-modality approaches.  Several works~\cite{linguraru_novel_2024}\cite{linguraru_lga_2024}\cite{li_lvit_2023} have introduced different effective text-image fusion strategies on datasets with slice-level textual annotations, such as QaTa-COV19 (X-ray)~\cite{degerli_osegnet_2022} and MosMedData+ (CT)~\cite{morozov_mosmeddata_2020}, significantly improving segmentation accuracy. However, in the field of brain tumor segmentation, no similar text-image dataset and text-image multimodal segmentation methods.

To address the aforementioned challenges, we develop TextBraTS, the first publicly available volume-level text-image brain tumor segmentation dataset. This dataset is based on the BraTS20 segmentation challenge training set~\cite{menze_multimodal_2014}, comprising 369 multimodal brain MRI scans, with expert textual annotations. Furthermore, we propose a volume-level text-guided image segmentation method, that utilizes bidirectional cross-attention mechanism to fuse high-level textual information with image features, thereby improving segmentation accuracy for multiple brain tumor regions. We also compare various template-based inputs for text reports and propose a template processing method for brain tumor reports. As shown in Fig.~\ref{overview}, compared to existing methods, this work provides a novel dataset and a foundational approach for volume-level text-guided medical image segmentation. 
Our key contributions are as follows:
\begin{enumerate}
\item We create the first publicly available volume-level text-image brain MRI tumor segmentation dataset, which includes multimodal MRI volumes and their corresponding detailed textual descriptions.
\item We propose a text-guided image segmentation model that incorporates a bidirectional cross-attention fusion mechanism, significantly improving segmentation accuracy.
\item Through extensive experiments, we explore the impact of different text template inputs and present comprehensive experimental results on our dataset.
\end{enumerate}

\section{TextBraTS Dataset Creation}
In this paper, we create the TextBraTS dataset, a novel volume-level text-image pair dataset. This dataset is an extension of the BraTS2020 segmentation challenge dataset \cite{menze_multimodal_2014}, comprising four imaging modalities and one textual modality.
\begin{figure}
    \centering
    \includegraphics[width=0.8\linewidth]{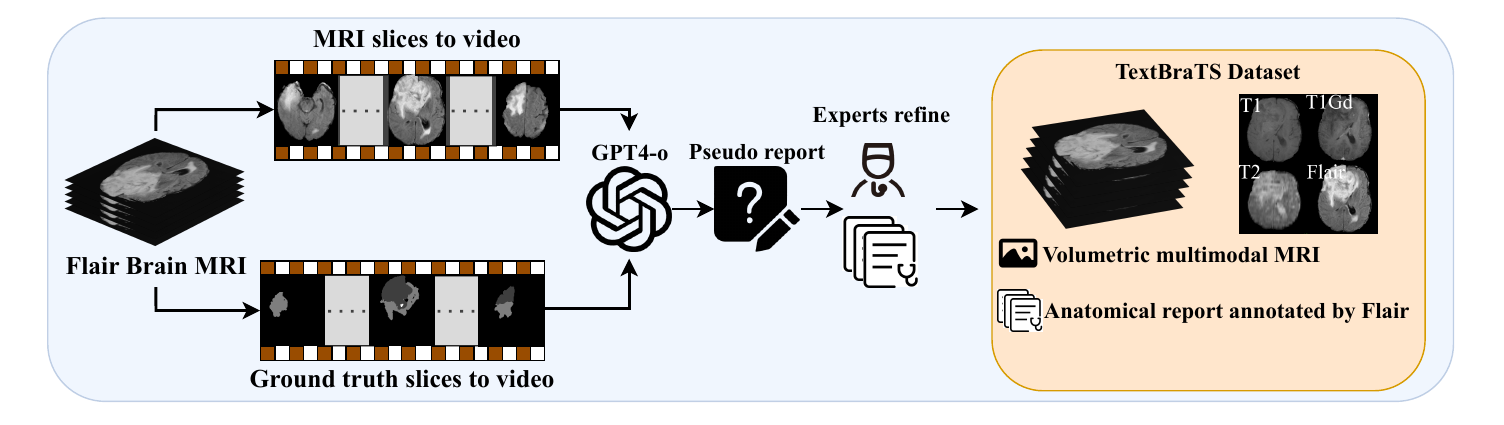}
    \caption{The TextBraTS dataset creation pipeline involves an initial annotation phase using the GPT-4o model to generate preliminary pseudo-reports. These reports are then refined by expert radiologists to ensure accuracy, resulting in the final TextBraTS dataset.}
    \label{pipeline}
\end{figure}
\subsubsection{Dataset Creation} 
The methodology for constructing the TextBraTS dataset is illustrated in Fig.~\ref{pipeline} . We created corresponding volume-level textual annotations for each case in the training set of the multimodal brain MRI dataset from the BraTS20 segmentation challenge \cite{menze_multimodal_2014}. Benefiting from the rapid advancements in multimodal large models, current models have demonstrated significant capabilities in analyzing and processing medical images. To reduce time costs, we leveraged the GPT-4o model \cite{openai_hello_2024} as a pre-annotation assistant. Each case in the BraTS20 segmentation challenge dataset includes four modalities: T1, T1Gd, T2, and Flair. According to the official dataset description \cite{menze_multimodal_2014}, tumor segmentation annotations are primarily based on the Flair modality. Therefore, our textual descriptions are also derived from the Flair modality. For each case, we sliced the 3D Flair images into individual frames and converted them into video format. Similarly, the corresponding tumor ground truth labels were sliced and converted into videos. These were then input into the GPT-4o model, guided by a carefully designed template prompt to generate analyses of the images. The analysis is divided into four parts: Overall lesion location and signal characteristics; Edema regions and features; Necrotic regions and features; Descriptions of midline shift and ventricular compression caused by the tumor's impact. Subsequently, experts reviewed the generated textual reports based on the images and segmentation labels, correcting the content to produce a final overall description.
\begin{figure*}[!t]
    \centering
    \includegraphics[width=0.7\linewidth]{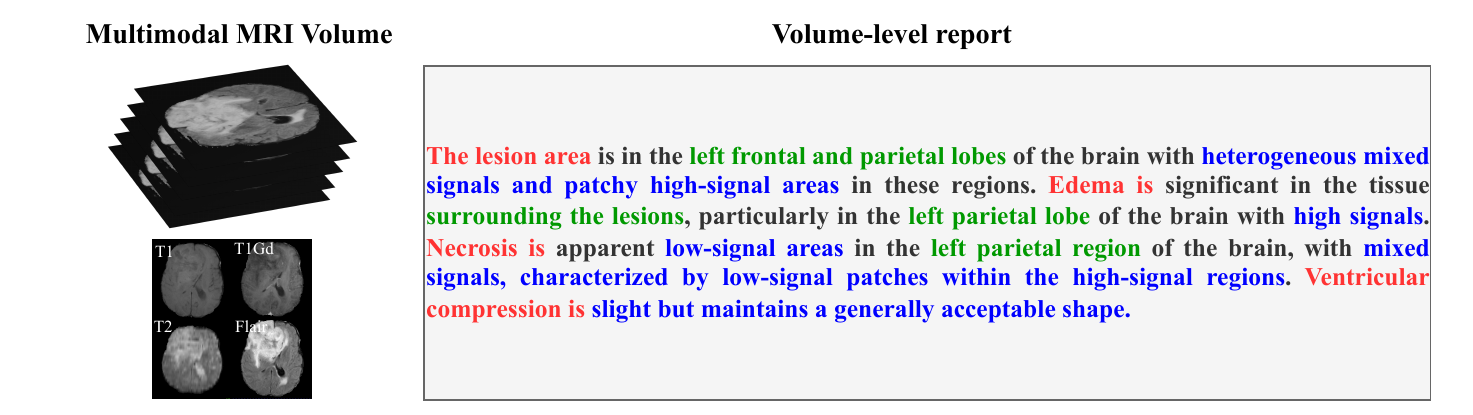}
    \caption{A sample in the TextBraTS dataset contains MRI data from four modalities and one text modality. The red part of the report represents the template words, the green part represents the location descriptions, and the blue part represents the features.}
    \label{data}
\end{figure*}
\subsubsection{Dataset Analysis}
Compared to existing medical text report datasets \cite{degerli_osegnet_2022}\cite{morozov_mosmeddata_2020}, the key distinction of our dataset is featured by comprehensive anatomical information and radiological description. The locations are described using the 3D spatial coordinates of the brain, such as the frontal lobe, temporal lobe, and other anatomical structures. The incorporated features include details such as signal intensities, the presence of discrete abnormal signal clusters, and other radiological characteristics. We adopted a simplified, template-based approach for the reports, guided by our experimental results. The details of this approach will be discussed in the Experiments section. An example of the TextBraTS textual annotations is shown in Fig.~\ref{data}.
\section{Text-Guided Volumetric Brain Tumor Segmentation}

 Based on the TextBraTS dataset, we propose a text-guided integration segmentation model that effectively improves segmentation performance. Additionally, our approach provides foundational strategies to guide the fusion of 3D imaging and textual information for related tasks.
\subsubsection{Overview of the model}
\begin{figure*}[!t]
    \centering
    \includegraphics[width=0.8\linewidth]{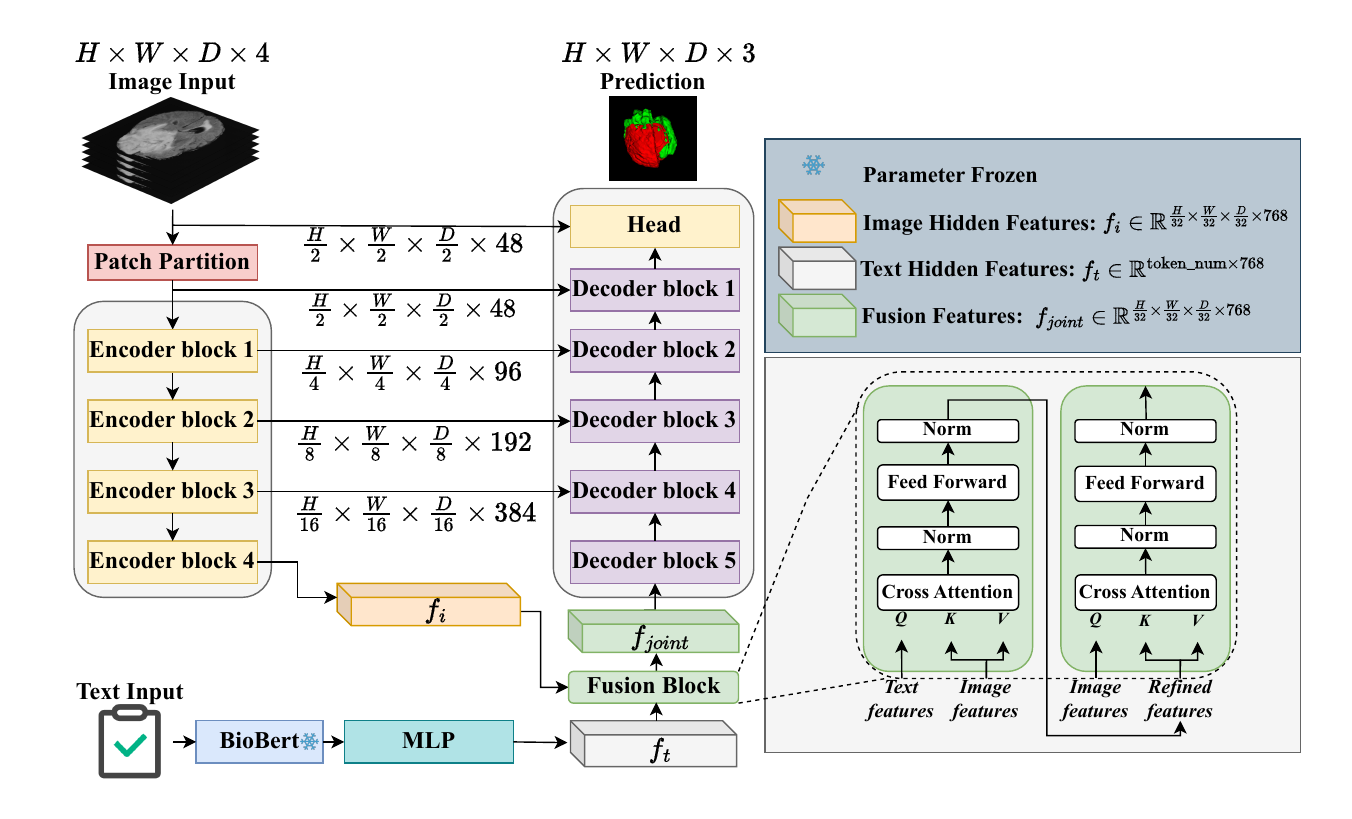}
    \caption{An overview of our text-guided brain tumor segmentation approach is presented. Our approach is built on a Transformer-based segmentation model as the backbone. To effectively integrate text and image information, we introduce a bidirectional cross-attention fusion block, enabling enhanced feature interaction between the two modalities.  }
    \label{method}
\end{figure*}
The structure of our framework is illustrated in Fig.~\ref{method}. We utilize the pre-trained language model BioBERT \cite{lee_biobert_2020} to extract text embeddings. BioBERT is pretrained on a vast amount of biomedical knowledge giving it strong feature extraction capabilities for medical reports. Since BioBERT is based on a Transformer architecture, we adopt SwinUNETR \cite{crimi_swin_2022},  also a Transformer-based model, as our image feature extractor, incorporating an encoder and decoder design.  This architectural alignment facilitates better fusion of text and image modalities by ensuring compatibility in feature map sizes. First, the model extracts text features using the frozen BioBERT, then employs a multi-layer perceptron (MLP) \cite{popescu_multilayer_2009} to map the text features into the image feature space for alignment. This process is described by the following formula:
\begin{equation}
f_t=MLP(BioBERT(t)).
\end{equation}
Let \(t\) be the text input, and \(f_t \in \mathbb{R}^{\text{token\_num} \times 768}\) is the text features. The token number is set as 110, based on the average length of text data. The output features of the image encoder can be described as \(f_i \in \mathbb{R}^{\frac{H}{32} \times \frac{W}{32} \times \frac{D}{32} \times 768}\), where the \(H,W,D\) are the input size of image encoder. Then, the \(f_t\) and \(f_i\) are sent to fusion block:
\begin{equation}
f_{joint}=Fusion(f_t,f_i).
\end{equation}
Finally, the fused features \(f_{joint}\in \mathbb{R}^{\frac{H}{32} \times \frac{W}{32} \times \frac{D}{32} \times 768}\), which maintain the same size as \(f_i\) are passed to the decoder to complete the segmentation of the target regions.
\subsubsection{Bidirectional cross-attention fusion}
For the fusion strategy, we propose a bidirectional cross-attention method. To further align the text and image modalities, we first use text information to refine the image features. The \(f_i\)  is reshaped as \(f_i \in \mathbb{R}^{\frac{H}{32}  \frac{W}{32}  \frac{D}{32} \times 768}\). The first cross-attention triplet of (Query, Key, Value) is computed as:
\begin{equation}
Q=f_tW_q,\space K=f_iW_k, \space V=f_iW_v.
\end{equation}
The text-guided refined image features \(f_i^{\prime}\) is computed as:
\begin{equation}
f_i^{\prime}=M((Softmax(\frac{QK^{T}}{\sqrt{d}})V),
\end{equation}
where the \(M\) is a module consisting of two normalization layers with a linear layer in between.
Through the first cross-attention mechanism, the refined features \(f_i^{\prime} \in \mathbb{R}^{\text{token\_num} \times 768}\)  are updated to capture shared information from text modality. These refined features are used to guide the original features, further improving the segmentation performance. The second cross-attention can be expressed as:
\begin{equation}
Q^{\prime}=f_iW_q^{\prime},\space K^{\prime}=f_i^{\prime}W_k^{\prime}, \space V^{\prime}=f_i^{\prime}W_v^{\prime},
\end{equation}
\begin{equation}
f_{joint}=M^{\prime}((Softmax(\frac{Q^{\prime}K^{\prime T}}{\sqrt{d}})V^{\prime}),
\end{equation}
where the \(M^{\prime}\) has the same structure as \(M\). The text-guided image features \(f_{joint}\) are sent to the first decoder layer for segmentation prediction.

\section{Experiment}
\subsubsection{Dataset and Evaluation Metrics}
The experimental dataset used in this study is the TextBraTS dataset proposed in this paper. The dataset consists of 369 brain MRI cases with brain tumors, where each case includes four imaging modalities (T1, T1Gd, T2, and Flair) and one text modality (radiology report). The segmentation task focuses on three regions: whole tumor (WT), enhancing tumor (ET), and tumor core (TC). The dataset was randomly split into training (220 cases), validation (55 cases), and testing (94 cases) sets. During testing, the checkpoint with the best performance on the validation set was used.The evaluation metrics used in this study are the Dice Similarity Coefficient (Dice)~\cite{dice_measures_1945} and Hausdorff Distance at the 95th Percentile (Hd95) \cite{huttenlocher_comparing_1993}.
\subsubsection{Implementation Details}
Our text-guided Brain Tumor Segmentation model was implemented using PyTorch \cite{paszke_automatic_2017} and MONAI \cite{cardoso_monai_2022} and trained on two NVIDIA RTX A6000 GPUs. We used a batch size of 2 per GPU, with a learning rate of 0.0001, and the optimizer was AdamW \cite{loshchilov_decoupled_2017}. For initialization, the weights of the SwinUNETR Transformer encoder were loaded from NVIDIA's providing model. A warm-up training strategy was applied for the first 50 epochs, with a total of 200 epochs of training. Since text data cannot undergo the same augmentation and cropping operations as images, we resampled all data to a fixed size of 128×128×128 for both training and testing, ensuring consistency across modalities.

\begin{table}[t]
\centering
\caption{Comparison of Segmentation Performance (Dice and HD95 Metrics)}
\renewcommand{\arraystretch}{1} % Adjust row spacing
\setlength{\tabcolsep}{8pt} % Adjust column spacing
\resizebox{\textwidth}{!}{%
\begin{tabular}{lcccccccc}
\toprule
\textbf{Method} & \multicolumn{4}{c}{\textbf{Dice (\%) $\uparrow$} 
} & \multicolumn{4}{c}{\textbf{HD95 $\downarrow$}} \\
\cmidrule(lr){2-5} \cmidrule(lr){6-9}
 & ET & WT & TC & Avg. & ET & WT & TC & Avg. \\
\midrule
3D-UNet \cite{navab_u-net_2015}  & 80.4 &87.3  &81.6  & 83.1 & 6.11 & 10.51  &8.93  & 8.17  \\
nnU-Net \cite{isensee_nnu-net_2021} & 82.2 & 87.5 & 82.6 & 84.1 & \textbf{4.27} & 11.90 & 8.52 & 8.23 \\
SegResNet \cite{crimi_brain_2022}& 80.9 & 88.4 & 82.3 & 83.8 & 6.18 & 7.28 & 7.41 & 6.95 \\
SwinUNETR \cite{crimi_swin_2022}& 81.0 & 89.5 & 80.8 & 83.8 & 5.95 & 8.23 & 7.03 & 7.07 \\
Nestedformer \cite{wang_nestedformer_2022}&82.6  & 89.5 &80.2  &84.1  &5.08  &10.51  &8.93  & 8.17  \\
Our work & \textbf{83.3} & \textbf{89.9} & \textbf{82.8} & \textbf{85.3} & 4.58 & \textbf{5.48} & \textbf{5.34} & \textbf{5.13} \\
\bottomrule
\end{tabular}
}
\label{sota}
\end{table}
\subsubsection{Comparison with SOTA methods}
\begin{figure}
    \centering
    \includegraphics[width=0.55\linewidth]{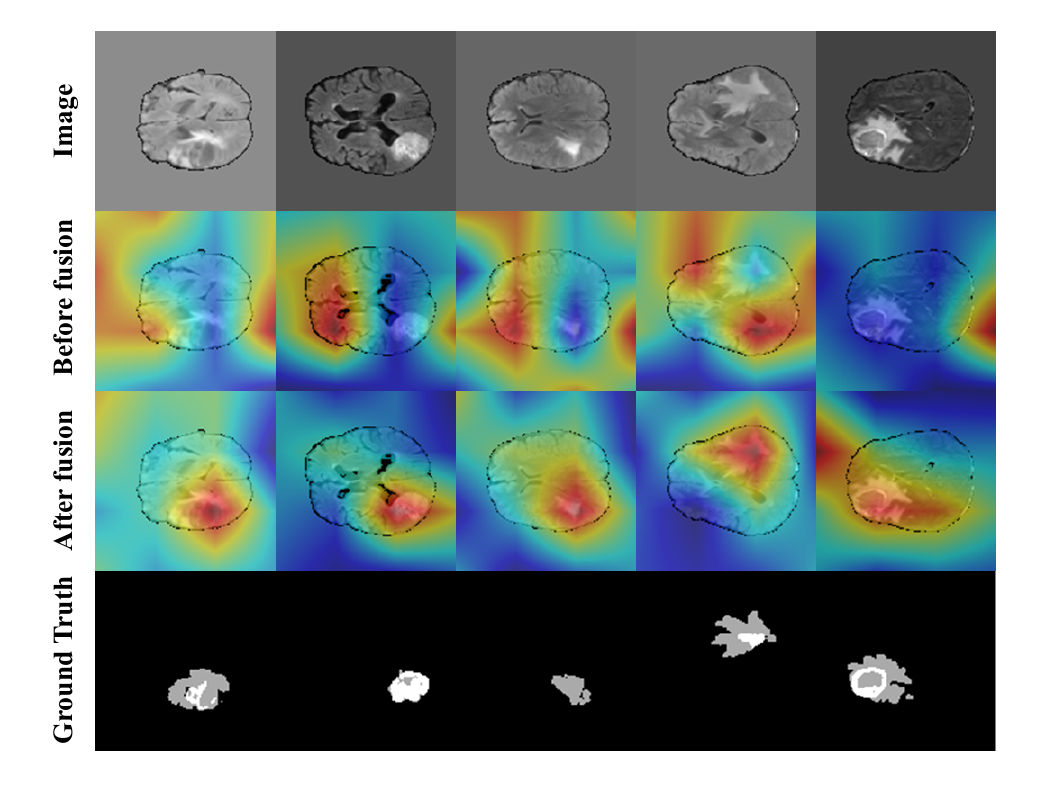}
    \caption{Visualization of features before fusion block and after fusion block }
    \label{visual}
\end{figure}
We compared several state-of-the-art (SOTA) methods based on CNN and Transformer architectures. The results, as shown in Table~\ref{sota}, demonstrate that our text-guided image segmentation model, developed using the newly proposed TextBraTS dataset, outperforms existing SOTA methods across all segmentation regions. To evaluate the performance difference between our model and SOTA methods \cite{wang_nestedformer_2022}, we conducted 10 independent training and testing runs. A t-test revealed a statistically significant difference with a p-value of 0.0077 (< 0.05). We visualized the feature maps before and after the fusion module, with some slice examples shown in Fig.~\ref{visual}. In some cases, it can be observed that the image features guided by text exhibit a significantly improved perception of tumor regions. This indicates that the text effectively assists in image segmentation.

\begin{table}[t]
\centering
\caption{Ablaion study on different input text reports}
\renewcommand{\arraystretch}{1}
\begin{tabular}{lcccccccccc}
\hline
 & \textbf{Location} & \textbf{Feature} & \multicolumn{4}{c}{\textbf{Dice (\%) $\uparrow$}} & \multicolumn{4}{c}{\textbf{HD95 $\downarrow$}} \\
\cline{4-11}
 & & &\textbf{ET} & \textbf{WT} & \textbf{TC} & \textbf{Avg.} & \textbf{ET} & \textbf{WT} & \textbf{TC} & \textbf{Avg.} \\
\hline
 \textbf{Raw text} & \checkmark & \checkmark & 83.0 & 89.6 & 82.1 & 84.6 & 5.96 & 7.11 & 6.32 & 6.46 \\
 \hline
  \multirow{3}{*}{\textbf{Templated}} & \checkmark & & 82.5 & 89.9 & 81.5 & 84.6 & 5.94 & 7.21 & 7.28 & 6.81 \\
 & &  \checkmark & 82.2 & 89.6 & 82.0 & 84.6 & 5.25 & 5.83 & 6.30 & 5.78 \\
 & \checkmark & $\checkmark$ & \textbf{83.3} & \textbf{89.9} & \textbf{82.8} & \textbf{85.3} & \textbf{4.58} & \textbf{5.48} & \textbf{5.34} & \textbf{5.13} \\
\hline
\end{tabular}
\label{template}
\end{table}
\begin{table}[t]
\centering
\caption{Ablation study of different fusion modules}
\renewcommand{\arraystretch}{1}
\begin{tabular}{lccccccccccc}
\hline
\textbf{Method} & \textbf{T2I} & \textbf{I2T} & \multicolumn{4}{c}{\textbf{Dice (\%) $\uparrow$}} & \multicolumn{4}{c}{\textbf{HD95 $\downarrow$}} \\
\cline{4-11}
& & & \textbf{ET} & \textbf{WT} & \textbf{TC} & \textbf{Avg.} & \textbf{ET} & \textbf{WT} & \textbf{TC} & \textbf{Avg.} \\
\hline
Dot Sum & & & 81.6 & 87.0 & 78.8 & 81.6 & 5.96 & 11.90 & 7.41 & 8.57 \\
Cross-attention & $\checkmark$ & & 82.5 & 89.9 & 82.1 & 84.8 & 5.43 & 7.18 & 6.18 & 6.26 \\
Bidirectional cross-attention & $\checkmark$ & $\checkmark$ & \textbf{83.3} & \textbf{89.9} & \textbf{82.8} & \textbf{85.3} & \textbf{4.58} & \textbf{5.48} & \textbf{5.34} & \textbf{5.13} \\
\hline
\end{tabular}
\label{fusion}
\end{table}

\subsubsection{Ablation study}
Excessive and highly diverse text descriptions in small-scale medical image datasets may hinder model generalization. To address this, we explored different text formats and content representations. First, we standardized a template format to enhance consistency. Then, we evaluated four types of text inputs: raw text, location-only templated, features-only templated, and fully templated, to determine the most effective approach for guiding segmentation. The results, presented in Table \ref{template}, show that fully templated text inputs significantly improve segmentation performance for both the tumor interior and its edges. By analyzing the results, we observed that location-only inputs achieved better overall tumor segmentation, as indicated by higher Dice scores. In contrast, feature-only inputs enhanced edge segmentation accuracy, as reflected in lower HD95 values.  We believe this is because location information is more effective for identifying the overall tumor position, whereas feature information provides a detailed description of abnormal signals, which better aids edge segmentation. These results suggest that structured and templated text inputs play a crucial role in optimizing segmentation performance by providing more consistent and interpretable guidance for the model.
The ablation study for the fusion module presented in Table~\ref{fusion}, compares three different methods: the dot sum of two modalities, one-step cross-attention for image features, and bidirectional cross-attention (T2I: Text to Image and I2T: Image to Text). The results demonstrate that the bidirectional cross-attention approach achieves the best performance, highlighting its effectiveness in integrating text and image features for improved segmentation accuracy. Through the processing of bidirectional cross-attention, the two modalities are better aligned in the same space, leading to superior results compared to using cross-attention only once.
\section{Conclusion}
We create \textbf{TextBraTS}, a novel text-image brain tumor segmentation dataset, and a text-guided segmentation method using a bidirectional attention mechanism, which outperforms state-of-the-art approaches. While the combined use of text and imaging is crucial in clinical diagnosis, existing methods often overlook text due to limitations. Our dataset and method fill this gap, providing a structured solution for future research in multimodal medical diagnostics. Future work will explore advanced fusion techniques and develop more effective multimodal segmentation models.

\bibliographystyle{splncs04}
\bibliography{miccai}

\end{document}